\documentclass[acmtog,nonacm]{acmart}

\AtBeginDocument{%
  \providecommand\BibTeX{{%
    \normalfont B\kern-0.5em{\scshape i\kern-0.25em b}\kern-0.8em\TeX}}}

\acmYear{2025} 

\usepackage{enumitem}
\usepackage{graphicx}
\usepackage{subcaption}
\usepackage{booktabs}
\usepackage{multirow}
\usepackage{xspace}

\newcommand{\methodname}{DaS\xspace}

\begin{document}

\title{Diffusion as Shader: 3D-aware Video Diffusion for Versatile Video Generation Control}



\author{Zekai Gu}
\affiliation{
  \institution{Hong Kong University of Science and Technology}
  \city{Hong Kong}
  \country{China}
}
\author{Rui Yan}
\affiliation{
  \institution{Zhejiang University}
  \city{Hangzhou}
  \country{China}
}
\author{Jiahao Lu}
\affiliation{
  \institution{Hong Kong University of Science and Technology}
  \city{Hong Kong}
  \country{China}
}
\author{Peng Li}
\affiliation{
  \institution{Hong Kong University of Science and Technology}
  \city{Hong Kong}
  \country{China}
}
\author{Zhiyang Dou}
\affiliation{
  \institution{The University of Hong Kong}
  \city{Hong Kong}
  \country{China}
}
\author{Chenyang Si}
\affiliation{
  \institution{Nanyang Technological University}
  \city{Singapore}
  \country{Singapore}
}
\author{Zhen Dong}
\affiliation{
  \institution{Wuhan University}
  \city{Wuhan}
  \country{China}
}
\author{Qifeng Liu}
\affiliation{
  \institution{Hong Kong University of Science and Technology}
  \city{Hong Kong}
  \country{China}
}
\author{Cheng Lin}
\affiliation{
  \institution{The University of Hong Kong}
  \city{Hong Kong}
  \country{China}
}
\author{Ziwei Liu}
\affiliation{
  \institution{Nanyang Technological University}
  \city{Singapore}
  \country{Singapore}
}
\author{Wenping Wang}
\affiliation{%
  \institution{Texas A\&M University}
  \city{Texas}
  \country{U.S.A}
}
\author{Yuan Liu}
\affiliation{
  \institution{Hong Kong University of Science and Technology}
  \city{Hong Kong}
  \country{China}
}
\makeatletter
\let\@authorsaddresses\@empty
\makeatother
\renewcommand{\shortauthors}{Zekai Gu, et al.}

\begin{abstract}
Diffusion models have demonstrated impressive performance in generating high-quality videos from text prompts or images. 
However, precise control over the video generation process—such as camera manipulation or content editing—remains a significant challenge. 
Existing methods for controlled video generation are typically limited to a single control type, lacking the flexibility to handle diverse control demands.
In this paper, we introduce Diffusion as Shader (\methodname), a novel approach that supports multiple video control tasks within a unified architecture. 
Our key insight is that achieving versatile video control necessitates leveraging 3D control signals, as videos are fundamentally 2D renderings of dynamic 3D content. 
Unlike prior methods limited to 2D control signals, \methodname leverages 3D tracking videos as control inputs, making the video diffusion process inherently 3D-aware. This innovation allows \methodname to achieve a wide range of video controls by simply manipulating the 3D tracking videos. 
A further advantage of using 3D tracking videos is their ability to effectively link frames, significantly enhancing the temporal consistency of the generated videos.
With just 3 days of fine-tuning on 8 H800 GPUs using less than 10k videos, \methodname demonstrates strong control capabilities across diverse tasks, including mesh-to-video generation, camera control, motion transfer, and object manipulation. Codes and more results are available at \href{https://igl-hkust.github.io/das/}{https://igl-hkust.github.io/das/}.
\end{abstract}


\begin{teaserfigure}
\centering
  \includegraphics[width=\linewidth]{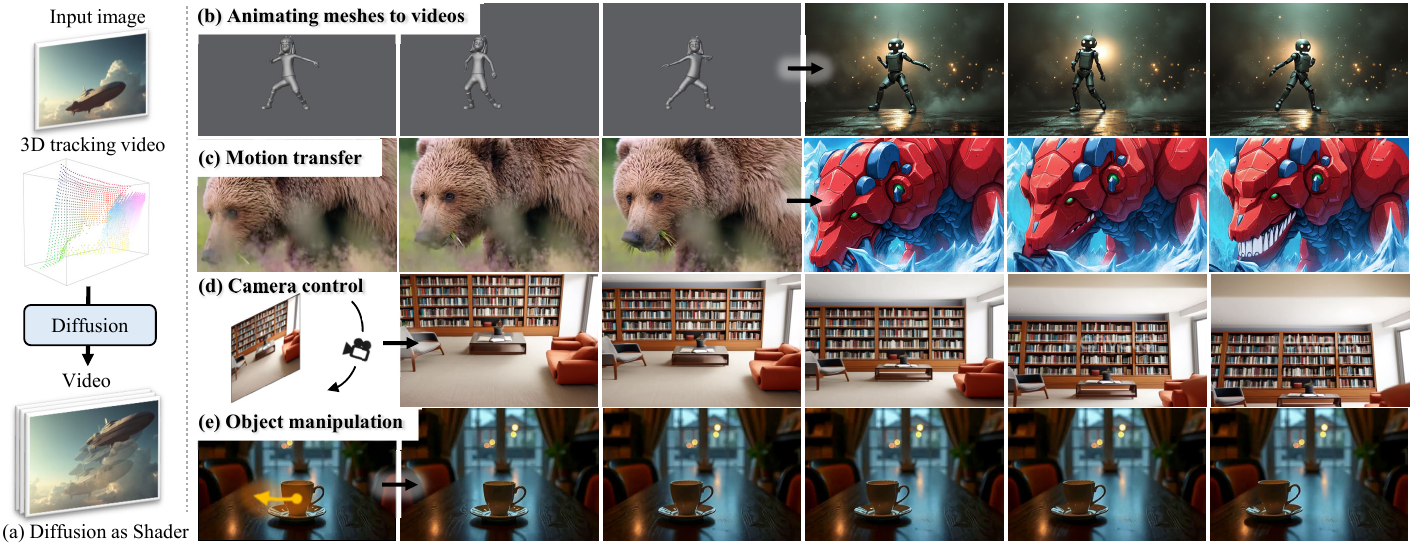}
  \vspace{-20pt}
  \caption{\textbf{Diffusion as Shader (\methodname)} is (a) a 3D-aware video diffusion method enabling versatile video control tasks including (b) animating meshes to video generation, (c) motion transfer, (d) camera control, and (e) object manipulation.}
  \label{fig:teaser}
\end{teaserfigure}

\maketitle
\vspace{-12pt}
\section{Introduction}

The development of diffusion generative models~\cite{rombach2022high,ho2020denoising,blattmann2023stable,brooks2024video,lin2024open,opensora} enables high-quality video generation from text prompts or a starting image. Recent emerging models, e.g. Sora~\cite{brooks2024video}, CogVideo-X~\cite{yang2024cogvideox}, Keling~\cite{keling}, and Hunyuan~\cite{kong2024hunyuanvideo}, have shown impressive video generation ability with strong temporal consistency and appealing visual effects, which becomes a promising tool for artists to create stunning videos using just few images or text prompts. These advancements show strong potential to revolutionize the advertising, film, robotics, and game industries, becoming fundamental elements for various generative AI-based applications.

A major challenge in video generation lies in achieving versatile and precise control to align seamlessly with users' creative visions. While recent methods have introduced strategies to integrate control into the video generation process~\cite{wang2024motionctrl,he2024cameractrl,polyak2024movie,he2024id,yuan2024identity,wang2024boximator,huang2023fine,guo2024sparsectrl,namekata2024sg,ma2024trailblazer,ma2024follow}, they predominantly focus on specific control types, relying on specialized architectures that lack adaptability to emerging control requirements. Furthermore, these approaches are generally limited to high-level adjustments—such as camera movements or maintaining identity—falling short when it comes to enabling fine-grained modifications, like precisely raising an avatar's left hand.

We argue that achieving versatile and precise video generation control fundamentally requires 3D control signals in the diffusion model. 
Videos are 2D renderings of dynamic 3D content. 
In a traditional Computer Graphics (CG)- based video-making pipeline, we can effectively control all aspects of a video in detail by manipulating the underlying 3D representations, such as meshes or particles. 
However, existing video control methods solely apply 2D control signals on rendered pixels, lacking the 3D awareness in the video generation process and thus struggling to achieve versatile and fine-grained controls.
Thus, to this end, we present a novel 3D-aware video diffusion method, called \textbf{Diffusion as Shader} (\methodname) in this paper, which utilizes 3D control signals to enable diverse and precise control tasks within a unified architecture. 

Specifically, as shown in Figure~\ref{fig:teaser} (a), \methodname is an image-to-video diffusion model that takes a 3D tracking video as the 3D control signals for various control tasks. 
The 3D tracking video contains the motion trajectories of 3D points whose colors are defined by their coordinates in the camera coordinate system of the first frame. In this way, the 3D tracking video represents the underlying 3D motion of this video. The video diffusion model acts like a shader to compute shaded appearances on the dynamic 3D points to generate the video. Thus, we call our model \textit{Diffusion as Shader}.

Using 3D tracking videos as control signals offers a significant advantage over depth videos with enhanced temporal consistency. While a straightforward approach to incorporating 3D control into video diffusion models involves using depth maps as control signals, depth maps only define the structural properties of the underlying 3D content without explicitly linking frames across time. In contrast, 3D tracking videos provide a consistent association between frames, as identical 3D points maintain the same colors across the video. These color anchors ensure consistent appearances for the same 3D points, thereby significantly improving temporal coherence in the generated videos. Our experiments demonstrate that even when a 3D region temporarily disappears and later reappears, \methodname effectively preserves the appearance consistency of that region, thanks to the temporal consistency enabled by the tracking video.

By leveraging 3D tracking videos, \methodname enables versatile video generation controls, encompassing but not limited to the following video control tasks.
\begin{enumerate}[itemsep=0pt,leftmargin=0.5cm]
    \vspace{-12pt}
    \item \textit{Animating meshes to videos}. Using advanced 3D tools like Blender, we can design animated 3D meshes based on predefined templates. These animated meshes are transformed into 3D tracking videos to guide high-quality video generation (Figure~\ref{fig:teaser} (b)).
    \item \textit{Motion transfer}. Starting with an input video, we employ a 3D tracker~\cite{xiao2024spatialtracker} to generate a corresponding 3D tracking video. Next, the depth-to-image Flux model~\cite{flux} is used to modify the style or content of the first frame. Based on the updated first frame and the 3D tracking video, \methodname generates a new video that replicates the motion patterns of the original while reflecting the new style or content (Figure~\ref{fig:teaser} (c)).
    \item \textit{Camera control}. To enable precise camera control, depth maps are estimated to extract 3D points~\cite{bochkovskii2024depth}. These 3D points are then projected onto a specified camera path to create a 3D tracking video, which guides the generation of videos with customized camera movements (Figure~\ref{fig:teaser} (d)).
    \item \textit{Object manipulation}. By integrating object segmentation techniques~\cite{kirillov2023segment} with a monocular depth estimator~\cite{bochkovskii2024depth}, the 3D points of specific objects can be extracted and manipulated. These modified 3D points are used to construct a 3D tracking video, which guides the creation of videos for object manipulation (Figure~\ref{fig:teaser} (e)).
\end{enumerate}



Due to the 3D awareness of \methodname, \methodname is data-efficient. Finetuning with less than 10k videos on 8 H800 GPUs for 3 days already gives the powerful control ability to \methodname, which is demonstrated by various control tasks. 
We compare \methodname with baseline methods on camera control~\cite{wang2024motionctrl,he2024cameractrl} and motion transfer~\cite{geyer2023tokenflow}, which demonstrates that \methodname achieves significantly improved performances in these two controlling tasks than baselines. 
For the remaining two tasks, i.e. mesh-to-video and object manipulation, we provide extensive qualitative results to show the superior generation quality of our method.

\section{Related Work}

\subsection{Video diffusion}

In recent years, the success of diffusion models in image generation~\cite{ho2020denoising, rombach2022high, peebles2023scalable} has sparked interest in exploring video generation~\cite{ho2022video, he2022latent, blattmann2023stable, chen2023videocrafter1, chen2024videocrafter2, brooks2024video, yang2024cogvideox, keling, kong2024hunyuanvideo, lin2024open, opensora, xing2025dynamicrafter,guo2023animatediff}. VDM~\cite{ho2022video} is the first work to explore the feasibility of diffusion in the field of video generation. SVD~\cite{blattmann2023stable} introduces a unified strategy for training a robust video generation model. Sora~\cite{brooks2024video}, through training on extensive video data, suggests that scaling video generation models is a promising path towards building general-purpose simulators of the physical world. CogVideo-X~\cite{yang2024cogvideox}, 
VideoCrafter~\cite{chen2023videocrafter1,chen2024videocrafter2}, DynamiCrafter~\cite{xing2025dynamicrafter},  Keling~\cite{keling}, and Hunyuan~\cite{kong2024hunyuanvideo} have demonstrated impressive video generation performance with strong temporal consistency.

\textbf{Controllable video generation}. Existing works still lack an effective way to control the generation process. 
There are many works~\cite{wang2024motionctrl,he2024cameractrl,polyak2024movie,he2024id,yuan2024identity,wang2024boximator,huang2023fine,guo2024sparsectrl,namekata2024sg,ma2024trailblazer,ma2024follow,yu2024viewcrafter,ma2024follow,qiu2024freetraj} that introduce a specific control signal in the video generation process which can only achieve one control type like identity preserving, camera control, and motion transfer. 
Our method is more versatile in various video control types by using a 3D-aware video generation with 3D tracking videos as conditions.


\subsection{Controlled video generation}
We review the following 4 types of controlled video generation.

\textbf{Animating meshes to videos.}
Animating meshes to videos aims to texture meshes. Several works~\cite{cao2023texfusion,richardson2023texture,wang2023breathing,cai2024generative} have demonstrated the feasibility of mesh texturization using powerful diffusion models.
TexFusion~\cite{cao2023texfusion} applies the diffusion model’s denoiser on a set of 2D renders of the 3D object, optimizing an intermediate neural color field to output final RGB textures. 
TEXTure~\cite{richardson2023texture} introduces a dynamic trimap representation and a novel diffusion sampling process, leveraging this trimap to generate seamless textures from various views. 
G-Rendering~\cite{cai2024generative} takes a dynamic mesh as input. To preserve consistency, G-Rendering employs UV-guided noise initialization and correspondence-aware blending of both pre- and post-attention features. Following G-Rendering, our method also targets dynamic meshes, utilizing a diffusion model as a shader to incorporate realistic texture information. Unlike G-Rendering, which preserves consistency at the noise and attention levels, our approach leverages 3D tracking videos as supplementary information, integrating them into the diffusion model to ensure both temporal and spatial consistency.

\textbf{Camera control.}
Camera control~\cite{wang2024motionctrl,he2024cameractrl,zheng2024cami2v,yu2024viewcrafter,bahmani2024ac3d,yang2024direct,xiao2024trajectory,geng2024motion,wang2024cpa} is an important capability for enhancing the realism of generated videos and increasing user engagement by allowing customized viewpoints. Recently, many efforts have been made to introduce camera control in video generation. MotionCtrl~\cite{wang2024motionctrl} incorporates a flexible motion controller for video generation, which can independently or jointly control camera motion and object motion in generated videos. CameraCtrl~\cite{he2024cameractrl} adopts Plücker embeddings~\cite{sitzmann2021light} as the primary form of camera parameters, enabling the ViewCrafter~\cite{yu2024viewcrafter} employs a point-based representation for free-view rendering, enabling precise camera control. AC3D~\cite{bahmani2024ac3d} optimizes pose conditioning schedules during training and testing to accelerate convergence and restricts the injection of camera conditioning to specific positions, reducing interference with other meaningful video features. 
CPA~\cite{wang2024cpa} incorporates a Sparse Motion Encoding Module to embed the camera pose information and integrating the embedded motion information via temporal attention.
Our method aims to use 3D tracking videos as an intermediary to achieve precise and consistent camera control.

\textbf{Motion transfer.}
Motion transfer~\cite{esser2023structure,geyer2023tokenflow,pondaven2024video,wang2024motionctrl,wang2024videocomposer,park2024spectral,yatim2024space,meral2024motionflow,geng2024motion} aims to synthesize novel videos by following the motion of the original one. Gen-1~\cite{esser2023structure} employs depth estimation results~\cite{ranftl2020towards, bochkovskii2024depth, lu2024align3r} to guide the motion. TokenFlow~\cite{geyer2023tokenflow} achieves consistent motion transfer by enforcing consistency in the diffusion feature space. MotionCtrl~\cite{wang2024motionctrl} also achieves motion transfer by incorporating a motion controller. DiTFlow~\cite{pondaven2024video} proposes Attention Motion Flow as guidance for motion transfer on DiTs~\cite{peebles2023scalable}. Motion Prompting~\cite{geng2024motion} utilizes 2D motions as prompts to realize impressive motion transfer. Unlike these approaches, our method employs 3D tracking as guidance for motion transfer, enabling a more comprehensive capture of each object's motion and the relationships between them within the video. This ensures accurate and globally consistent geometric and temporal consistency.

\begin{figure*}
    \centering
    \includegraphics[width=\textwidth]{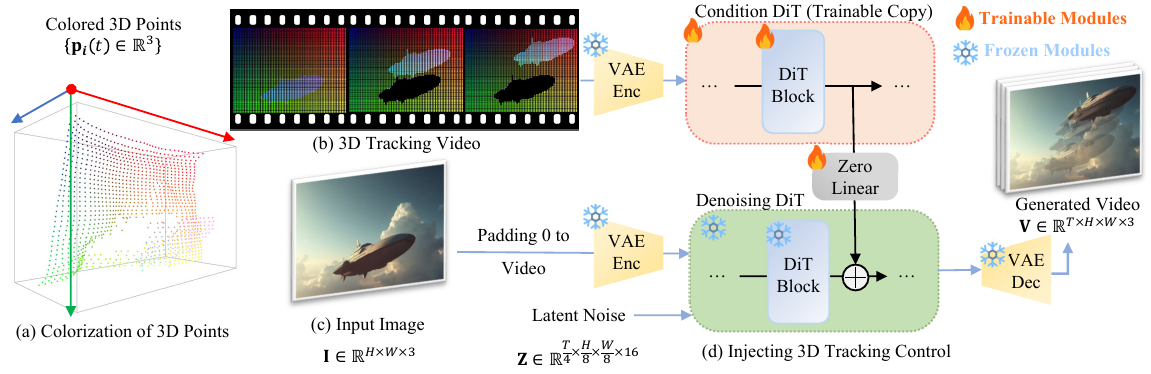}
    \caption{\textbf{Architecture of \methodname}. (a) We colorize dynamic 3D points according to their coordinates to get (b) a 3D tracking video. (c) The input image and the 3D tracking video are processed by (d) a transformer-based latent diffusion with a variational autoencoder (VAE). The 3D tracking video is processed by a trainable copy of the denoising DiT and zero linear layers are used to inject the condition features from 3D tracking videos into the denoising process.}
    \label{fig:pipe}
\end{figure*}

\textbf{Object manipulation.} 
Object manipulation refers to versatile object movement control for image-to-video generation. Different from camera control, which focuses on changes in perspective, object manipulation emphasizes the movement of the objects themselves. 
Currently, mainstream methods~\cite{chen2023motion,li2024image,mou2024revideo,teng2023drag,wang2024motionctrl,yin2023dragnuwa,jain2024peekaboo,ma2024trailblazer,qiu2024freetraj,wang2024boximator,yang2024direct,geng2024motion} typically achieve object manipulation by utilizing directed trajectories or modeling the relationships between bounding boxes with specific semantic meanings. However, these methods primarily rely on 2D guidance to represent the spatial movement of target objects, which often fails to accurately capture user intent and frequently results in distorted outputs. ObjCtrl-2.5D~\cite{wang2024objctrl} tries to address this limitation by extending 2D trajectories with depth information, creating a single 3D trajectory as the control signal. Better than the single 3D trajectory, our method leverages 3D tracking videos, which offer greater details and more effectively represent the motion relationships between foreground and background for more precise and realistic object manipulation.


\textbf{Concurrent works}. 
Recently, several works~\cite{geng2024motion, niu2025mofa, koroglu2024onlyflow, jeong2024track4gen, shi2024motion, lei2024animateanything,feng2024i2vcontrol,zhang2024world} have explored utilizing motion as control signals. These approaches can be broadly categorized into two groups: 2D motion-based and 3D motion-based methods.  
\cite{koroglu2024onlyflow, shi2024motion,lei2024animateanything} leverage 2D optical flow to condition motion, while \cite{niu2025mofa,geng2024motion,jeong2024track4gen} utilize 2D tracks, which are sparser than optical flow, to track or control video motion. 
\cite{zhang2024world} learns to generate 3D coordinates in the video diffusion model, which 3D awareness. \cite{feng2024i2vcontrol} lifts videos into 3D space and extracts the motion of 3D points, enabling a more accurate capture of spatial relationships between objects and supporting tasks such as object manipulation and camera control. Our method, \methodname, also leverages recent tracking methods~\cite{xiao2024spatialtracker,zhang2024protracker} to construct videos. However, we extend the applicability by unifying a broader range of control tasks, including mesh-to-video generation and motion transfer.

\section{Method}

\subsection{Overview}
\methodname is an image-to-video (I2V) diffusion generative model, which applies both an input image and a 3D tracking video as conditions for controllable video generation. In the following, we first review the backend I2V video diffusion model in Sec.~\ref{sec:dit}. Then, we discuss the definition of the 3D tracking video and how to inject the 3D tracking video into the generation process as a condition in Sec.~\ref{sec:tracking}. Finally, in Sec.~\ref{sec:control}, we discuss how to apply \methodname in various types of video generation control.

\subsection{Backend video diffusion model}
\label{sec:dit}

\methodname is finetuned from the CogVideoX~\cite{yang2024cogvideox} model that is a transformer-based video diffusion model~\cite{peebles2023scalable} operating on a latent space. 
Specifically, as shown in Figure~\ref{fig:pipe} (d), we adopt the I2V CogVideoX model as the base model, which takes an image $\mathbf{I}\in \mathbb{R}^{H\times W \times 3}$ as input and generate a video $\mathbf{V} \in \mathbb{R}^{T\times H\times W\times 3}$. The generated video $\mathbf{V}$ has $T$ frames with the same image size of width $W$ height $H$ as the input image. The input image $\mathbf{I}$ is first padded with zeros to get an input condition video with the same size $T\times H\times W \times 3$ as the target video. Then, a VAE encoder is applied to the padded condition video to get a latent vector of size $\frac{T}{4}\times \frac{H}{8}\times \frac{W}{8}\times 16$, which is concatenated with a noise of the same size. A diffusion transformer (DiT) \cite{peebles2023scalablediffusionmodelstransformers} is iteratively used to denoise the noise latent for a predefined number of steps and the output denoised latent is processed by a VAE decoder to get the video $\mathbf{V}$. In the following, we discuss how to add a 3D tracking video as an additional condition on this base model.

\begin{figure*}
    \centering
    \includegraphics[width=\textwidth]{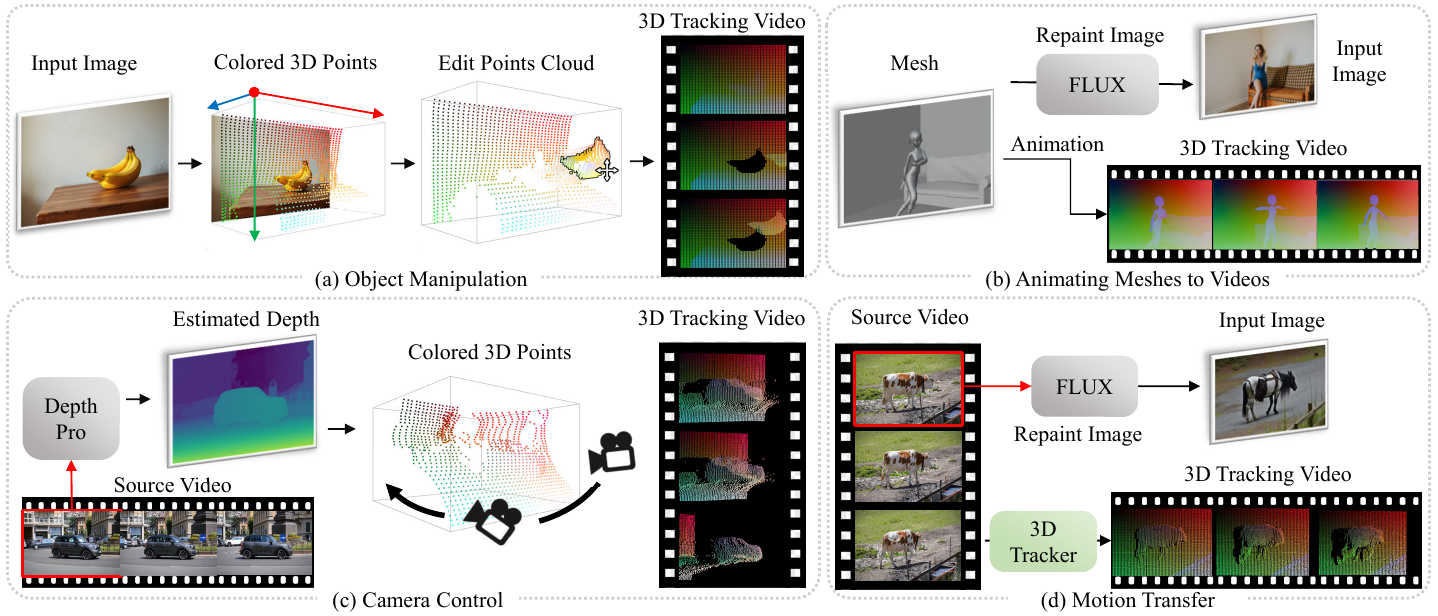}
    \caption{\textbf{3D tracking video generation} in (a) object manipulation, (b) animating mesh to video generation, (c) camera control, and (d) motion transfer.}
    \label{fig:pipe-tasks}
\end{figure*}
\subsection{Finetuning with 3D tracking videos}
\label{sec:tracking}

We add a 3D tracking video as an additional condition to our video diffusion model. As shown in Figure~\ref{fig:pipe} (a, b), the 3D tracking video is rendered from a set of moving 3D points $\{\mathbf{p}_i(t)\in\mathbb{R}^3\}$, where $t=1,...,T$ means the frame index in the video. The colors of these points are determined by their coordinates in the first frame, where we normalize the coordinates into $[0,1]^3$ and convert the coordinates into RGB colors $\{\mathbf{c}_i\}$. 
Note we adopt the reciprocal of z-coordinate in the normalization.
These colors remain the same for different timesteps $t$. Then, to get a specific $t$-th frame of the tracking video, we project these 3D points onto the $t$-th camera to render this frame. In Sec.~\ref{sec:control}, we will discuss how to get these moving 3D points and the camera poses of different frames for different control tasks. Next, we first introduce the architecture to utilize the 3D tracking video as a condition for video generation.

\textbf{Injecting 3D tracking control}. We follow a similar design as the ControlNet~\cite{zhang2023adding,chen2025pixart} in \methodname to add the 3D tracking video as the additional condition. As shown in Figure~\ref{fig:pipe} (d), we apply the pretrained VAE encoder to encode the 3D tracking video to get the latent vector. Then, we make a trainable copy of the pretrained denoising DiT, called condition DiT, to process the latent vector of the 3D tracking video. The denoising DiT contains 42 blocks and we copy the first 18 blocks as the condition DiT. In the condition DiT, we extract the output feature of each DiT block, process it with a zero-initialized linear layer, and add the feature to the corresponding feature map of the denoising DiT. We finetune the condition DiT with the diffusion losses while freezing the pretrained denoising DiT. 


\textbf{Finetuning details}.
To train the \methodname model, we construct a training dataset containing both real-world videos and synthetic rendered videos. The real-world videos are from MiraData~\cite{ju2024miradatalargescalevideodataset} while we use the meshes and motion sequences from Mixamo to render synthetic videos. All videos are center-cropped and resized to $720 \times 480$ resolution with 49 frames. 
We only finetune the copied condition DiT while freezing all the original denoising DiT.
To construct the 3D tracking video for the rendered videos, since we have access to the ground-truth 3D meshes and camera poses for the synthetic videos, we construct our 3D tracking videos directly using these dense ground-truth 3D points, which results in dense 3D point tracking. For real-world videos, we adopt SpatialTracker~\cite{xiao2024spatialtracker} to detect 3D points and their trajectories in the 3D space. Specifically, for each real-world video, we detect 4,900 3D evenly distributed points and track their trajectories. For training, we employ a learning rate of $1\times 10^{-4}$ \ using the AdamW optimizer. We train the model for 2000 steps using the gradient accumulation strategy to get an effective batch size of 64. The training takes 3 days on 8 H800 GPUs.

\subsection{Video generation control}
\label{sec:control}
In this section, we describe how to utilize \methodname for the following controllable video generation.

\subsubsection{Object manipulation}
\methodname can generate a video to manipulate a specific object. As shown in Figure~\ref{fig:pipe-tasks} (a), given an image, we estimate the depth map using Depth Pro~\cite{bochkovskii2024depth} or MoGE~\cite{wang2024mogeunlockingaccuratemonocular} and segment out the object using SAM~\cite{kirillov2023segment}. Then, we are able to manipulate the point cloud of the object to construct a 3D tracking video for object manipulation video generation.

\begin{figure*}[h]
    \centering
    \includegraphics[width=\textwidth]{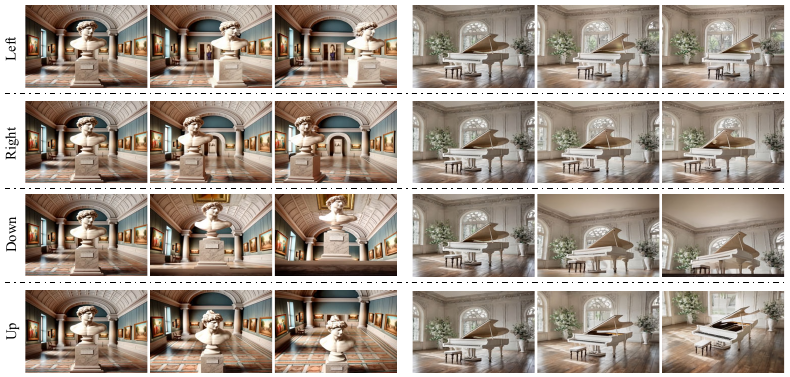}
    \caption{\textbf{Qualitative results of \methodname on the camera control task}. We show 4 trajectories (left, right, up, down) with large movements.}
    \label{fig:camctrl}
\end{figure*}

\subsubsection{Animating meshes to videos}
\methodname enables the creation of visually appealing, high-quality videos from simple animated meshes. While many Computer Graphics (CG) software tools provide basic 3D models and motion templates to generate animated meshes, these outputs are often simplistic and lack the detailed appearance and geometry needed for high-quality animations. Starting with these simple animated meshes, as shown in Figure~\ref{fig:pipe-tasks} (b), we generate an initial visually appealing frame using a depth-to-image FLUX model~\cite{flux}. We then produce 3D tracking videos from the animated meshes, which, when combined with the generated first frame, guide \methodname to transform the basic meshes into visually rich and appealing videos.


\subsubsection{Camera control}
Previous approaches~\cite{he2024cameractrl,wang2024motionctrl} rely on camera or ray embeddings as conditions to control the camera trajectory in video generation. However, these embeddings lack true 3D awareness, leaving the diffusion models to infer the scene's 3D structure and simulate camera movement. In contrast, \methodname significantly enhances 3D awareness by incorporating 3D tracking videos for precise camera control. To generate videos with a specific camera trajectory, as shown in Figure~\ref{fig:pipe-tasks} (c), we first estimate the depth map of the initial frame using Depth Pro~\cite{bochkovskii2024depth} and convert it into colored 3D points. These points are then projected onto the given camera trajectory, constructing a 3D tracking video that enables \methodname to control camera movements with high 3D accuracy.

\subsubsection{Motion transfer}
As shown in Figure~\ref{fig:pipe-tasks} (d), \methodname also facilitates creating a new video by transferring motion from an existing source video. First, we estimate the depth map of the source video’s first frame and apply the depth-to-image FLUX model~\cite{flux} to repaint the frame into a target appearance guided by text prompts. Then, using SpatialTracker~\cite{xiao2024spatialtracker}, we generate a 3D tracking video from the source video to serve as control signals. Finally, the \methodname model generates the target video by combining the edited first frame with the 3D tracking video.

\section{Experiments}


We conduct experiments on five tasks, including camera control, motion transfer, mesh-to-video generation, and object manipulation to demonstrate the versatility of \methodname in controlling the video generation process.


\subsection{Camera control}

\textbf{Baseline methods}. To evaluate the ability to control camera motions of generated videos, we select two representative methodologies, MotionCtrl~\cite{wang2024motionctrl} and CameraCtrl~\cite{he2024cameractrl} as baseline methods, both of which allow camera trajectories as input and use camera or ray embeddings for camera control.

\noindent\textbf{Metrics}. To measure the accuracy of the camera trajectories of generated videos, we evaluate the consistency between the estimated camera poses from the generated videos and the input ground-truth camera poses using rotation errors and translation errors.
Specifically, for each frame of a generated video, we reconstruct its relative pose given the first frame using SIFT~\cite{ng2003sift}. Then, we get the normalized quaternion and translation vectors for the rotation and translation. Finally, we calculate the cosine similarity between the estimated camera poses with the given camera poses.
\[
\mathbf{RotErr} = \arccos\left(\frac{1}{T-1} \sum_{i=2}^{T} \langle \;\mathbf{q}_{\mathrm{gen}}^i , \mathbf{q}_{\mathrm{gt}}^i \;\rangle\right),
\]
\[
\mathbf{TransErr} = \arccos \left(\frac{1}{T-1} \sum_{i=2}^{T} \langle\; \mathbf{t}_{\mathrm{gen}}^i , \mathbf{t}_{\mathrm{gt}}^i \;\rangle\right),
\]
where $T$ is the number of frames, $\mathbf{q}^{i}$ and $\mathbf{t}^{i}$ are the normalized quaternion and translation vector of the $i$-th frame, and $\langle \cdot, \cdot \rangle$ means the dot product between two vectors.

\noindent\textbf{Results}.
We compare against baseline methods on 100 random trajectories from RealEstate10K~\cite{zhou2018stereo}. But since most of the random trajectories only contain small movements, we further test the models on larger fixed movements (moving left, right, up, down, spiral) as shown in Figure~\ref{fig:camctrl}. As shown in Table~\ref{tab:camctrl}, our method outperforms the baseline methods, which demonstrates that our method achieves stable and accurate control of the camera poses of the generated videos. The main reason is that due to the utilization of the 3D tracking videos, our method is fully 3D-aware to enable accurate spatial inference in the video generation process. In comparison, baseline methods~\cite{he2024cameractrl,wang2024motionctrl} only adopt implicit camera or ray embeddings for camera control.

\begin{table}[]
\begin{tabular}{cllll}
\hline
Method     & \multicolumn{2}{c}{\textbf{Small Movement}} & \multicolumn{2}{c}{\textbf{Large Movement}} \\
           & TransErr $\downarrow$             & RotErr    $\downarrow$            & TransErr      $\downarrow$         & RotErr     $\downarrow$           \\ \hline
MotionCtrl & 44.23                  & 8.92                  & 67.05          & 39.86                  \\
CameraCtrl & 42.31                  & 7.82                  & 66.76                   & 29.70                 \\
Ours       & \textbf{27.85}         & \textbf{5.97}         & \textbf{37.17}                  & \textbf{10.40}         \\ \hline

\end{tabular}
\caption{\textbf{Quantitative results on camera control} of MotionCtrl~\cite{wang2024motionctrl}, CameraCtrl~\cite{he2024cameractrl}, and our method. ``TransErr'' and ``RotErr" are the angle differences between the estimated translation and rotation and the ground-truth ones in degree.}
\label{tab:camctrl}
\end{table}

\begin{figure*}[htbp]
    \centering
    \includegraphics[width=\textwidth]{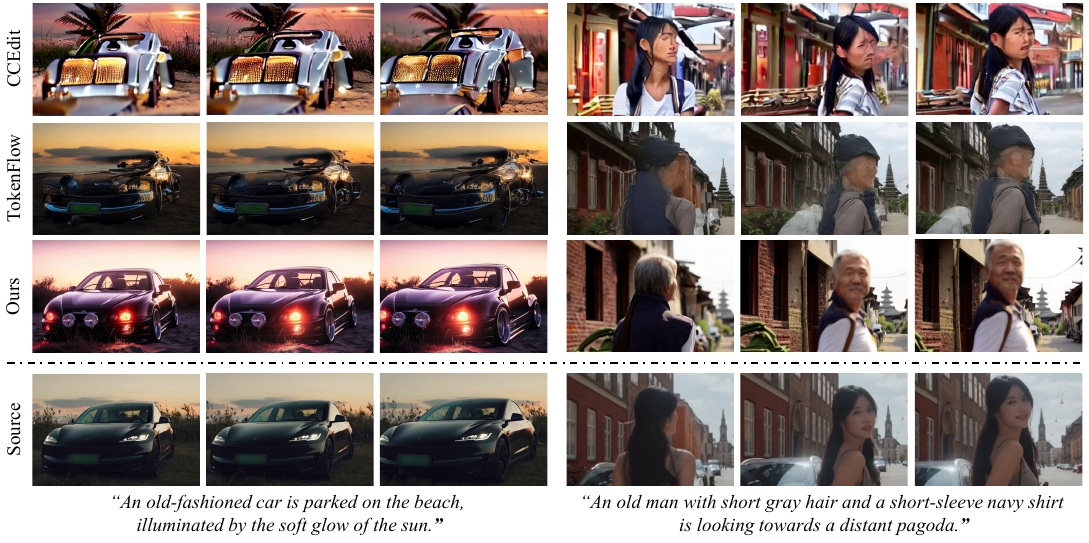}
    \caption{\textbf{Qualitative comparison on motion transfer} between our method, CCEdit~\cite{ccedit}, and TokenFlow~\cite{tokenflow}.}
    \label{fig:v2v2}
\end{figure*}

\subsection{Motion transfer}


\noindent\textbf{Baseline methods}. We compare \methodname with two famous motion transfer methods, TokenFlow~\cite{tokenflow} and CCEdit~\cite{ccedit}. TokenFlow represents video motions with the feature consistency across different timesteps extracted by a diffusion model. Then, the feature consistency is propagated to several keyframes generated by a text prompt for video generation. For TokenFlow, we adopt the Stable Diffusion 2.1~\cite{rombach2022high} model for the motion transfer task. CCEdit adopts depth maps as conditions to control the video motion and transfers the motion using a new repainted frame to generate a video. 

\noindent\textbf{Metrics}. 
Since all methods generate the transferred videos based on text prompts, we aim to evaluate the alignment between the generated videos and the text prompts, as well as the video coherence, using the CLIP \cite{radford2021learningtransferablevisualmodels}. Specifically, for video-text alignment, we extract multiple frames from the video and compare them with the corresponding text prompts by calculating the CLIP score \cite{hessel2022clipscorereferencefreeevaluationmetric} for each frame. This score reflects the alignment between image content and textual descriptions. For temporal consistency, we extract normalized CLIP features from adjacent video frames and compute the cosine similarity between the adjacent features. 

\begin{figure*}[htbp]
    \centering
    \includegraphics[width=\textwidth]{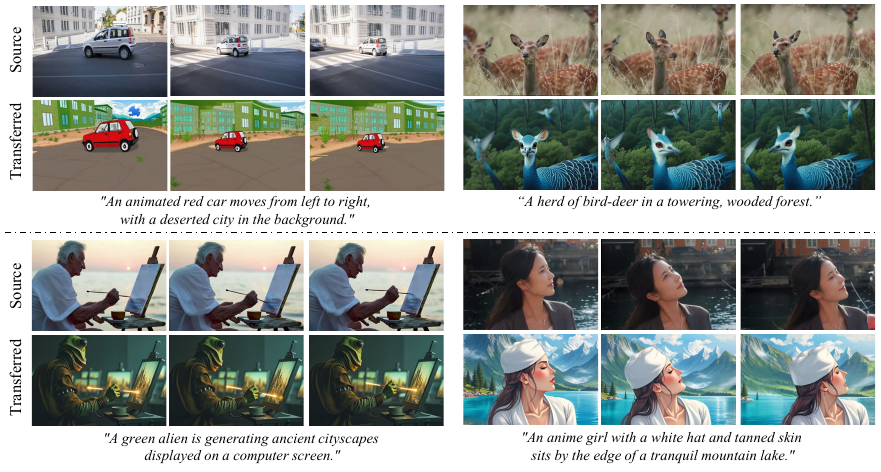}
    \caption{\textbf{Qualitative results on motion transfer of our method}.}
    \label{fig:v2v1}
\end{figure*}

\noindent\textbf{Results}. As shown in Table~\ref{tab:motion}, our method demonstrates outstanding performance in both text alignment and frame consistency, surpassing two baseline methods. Furthermore, \autoref{fig:v2v2} presents the qualitative comparison of our method, CCEdit, and TokenFlow. It shows that CCEdit produces frames of low quality and struggles to maintain temporal coherence. TokenFlow produces semantically consistent frames but has difficulty producing coherent videos. In contrast, our method accurately transfers the video motion with strong temporal coherence as shown in Figure~\ref{fig:v2v1}. 

\begin{table}[]
\centering
\begin{tabular}{lcc}
\hline
Method                           & \multicolumn{1}{c}{Tex-Ali} $\uparrow$ & \multicolumn{1}{c}{Tem-Con} $\uparrow$ \\ \hline
CCEdit                           & 16.9& 0.932 \\
Tokenflow                        & 31.9& 0.956 \\
Ours                             & \textbf{32.6}& \textbf{0.971} \\
\hline
\end{tabular}
\caption{\textbf{CLIP scores for motion transfer} of CCEdit~\cite{ccedit}, TokenFlow~\cite{tokenflow}, and our method. ``Text-Ali'' is the semantic CLIP consistency between generated videos and the given text prompts. ``Tem-Con'' is the temporal CLIP consistency between neighboring frames.}
\vspace{-15pt}
\label{tab:motion}
\end{table}

\subsection{Animating meshes to videos}
\begin{figure*}
    \centering
    \includegraphics[width=0.97\textwidth]{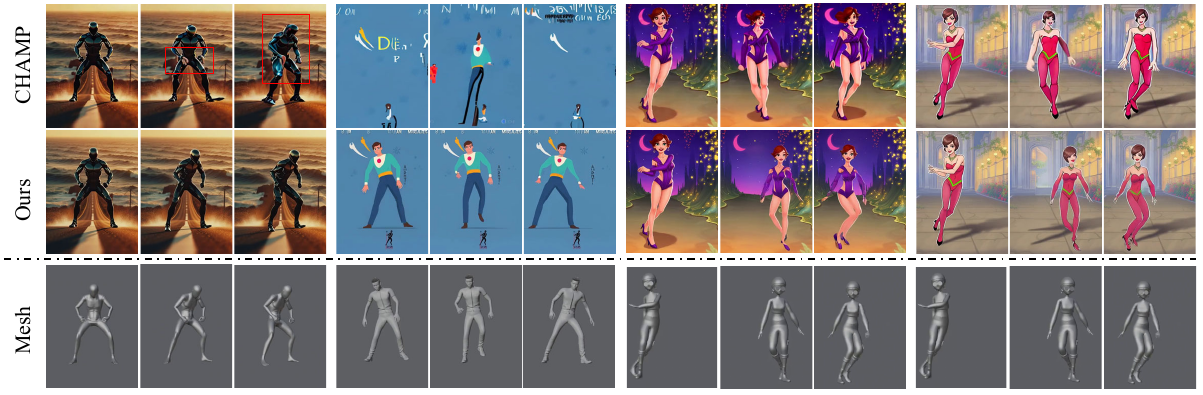}
    \caption{\textbf{Qualitative comparison on the animating mesh to video task} between our method and CHAMP~\cite{zhu2024champ}.}
    \label{fig:m2v2}
\end{figure*}

\textbf{Qualitative comparison}. We compare our method against a state-of-the-art human image animation method CHAMP \cite{zhu2024champ} on the mesh-to-video task. 
Champ takes a human image and a motion sequence as input and generates a corresponding human video. The motion sequence is represented by an animated SMPL~\cite{loper2023smpl} mesh. We use the same input image but the SMPL mesh for CHAMP and generate the corresponding animation videos for qualitative comparison as shown in \autoref{fig:m2v2}. We also generate different styles of videos from the same animated 3D meshes as shown in \autoref{fig:m2v2}.
Compared to CHAMP, our method demonstrates better consistency in the 3D structure and texture details of the avatar on different motion sequences and across different styles. 

\begin{figure*}
    \centering
    \includegraphics[width=\textwidth]{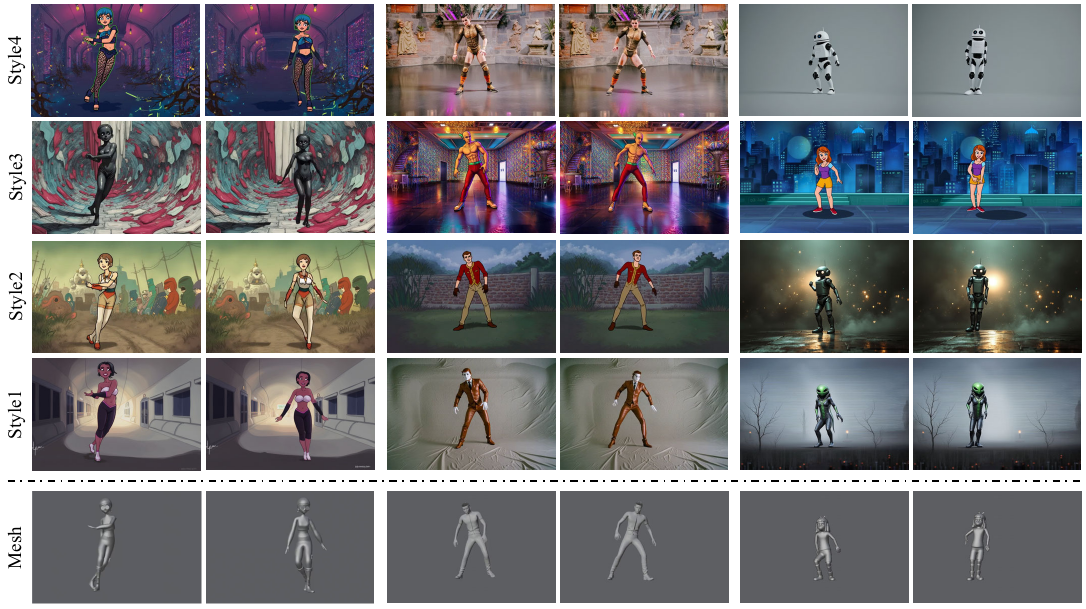}
    \caption{\textbf{More results of the animating mesh to video generation task}. Our method enables the generation of different styles from the same mesh.}
    \label{fig:m2v1}
\end{figure*}

\begin{figure*}
    \centering
    \includegraphics[width=\textwidth]{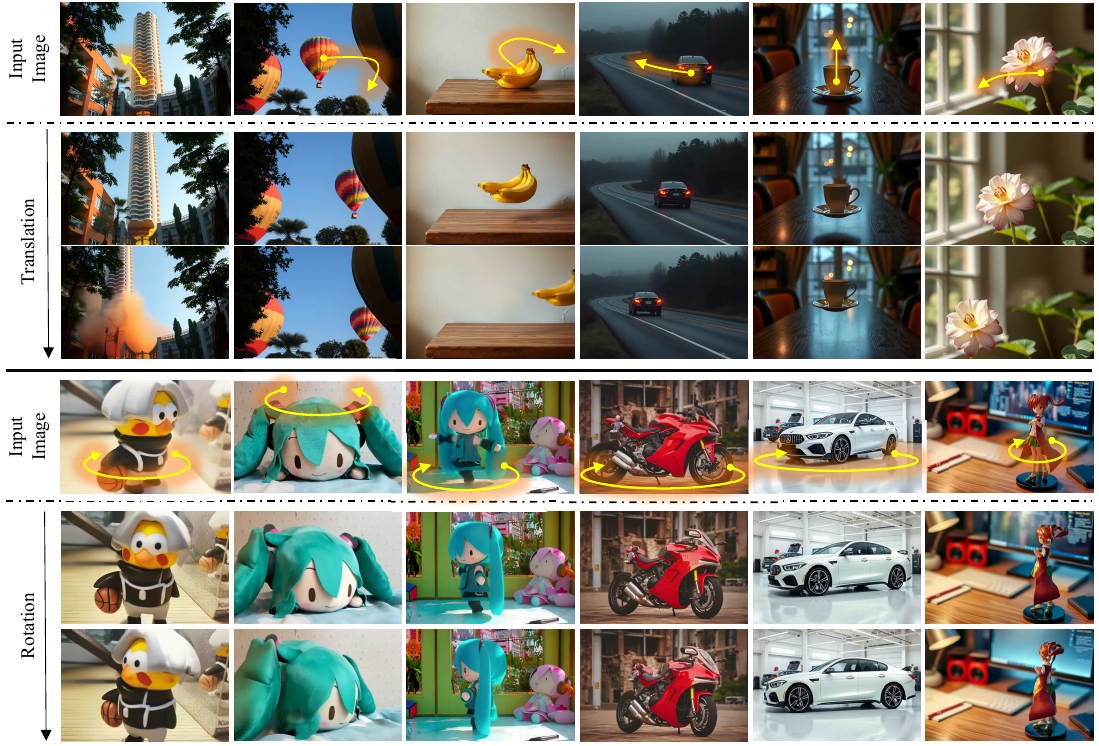}
    \caption{\textbf{Qualitative results of our method on the object manipulation task}. The top part shows the results of translation while the bottom part shows the results of rotating the object.}
    \label{fig:mani2d}
\end{figure*}

\subsection{Object manipulation}
\textbf{Qualitative results}. For the object manipulation, we adopt the SAM~\cite{kirillov2023segment} and depth estimation models~\cite{bochkovskii2024depth,wang2024mogeunlockingaccuratemonocular} to get the object points. Then, we evaluate two kinds of manipulation, i.e. translation and rotation. The results are shown in Figure~\ref{fig:mani2d}, which demonstrate that \methodname achieves accurate object manipulation to produce photorealistic videos with strong multiview consistency for these objects.

\begin{table}[]
\setlength\tabcolsep{3pt}
\centering
\begin{tabular}{ccccccc}
\hline
Depth & Tracking    & \#Tracks & PSNR $\uparrow$ & SSIM $\uparrow$ & LPIPS $\downarrow$ & FVD $\downarrow$ \\ \hline
\checkmark &  & -      & 18.08           & 0.573           & 0.312              & 645.1            \\
& \checkmark  & 900    & 18.52           & 0.586           & 0.337              & 765.3            \\
& \checkmark  & 2500   & 19.17           & 0.632           & 0.263              & 566.4            \\
& \checkmark  & 4900   & \textbf{19.27}  & \textbf{0.658}  & \textbf{0.261}     & \textbf{551.3}   \\
& \checkmark  & 8100   & 19.11           & 0.649           & 0.262              & 599.0            \\ \hline
\end{tabular}
\caption{\textbf{Analysis of applying different 3D control signals for image to video generation}. We evaluate PSNR, SSIM, LPIPS, and FVD of generated videos on the validation set of the DAVIS and MiraData datasets. ``Depth'' means using depth maps as the 3D control signals. ``Tracking'' means using 3D tracking videos as the control signals. \#Tracks means the number of 3D  points used in the 3D tracking video.}
\vspace{-15pt}
\label{tab:ablation}
\end{table}

\subsection{Analysis}

We conduct analysis on the choice of 3D control signals, i.e. depth maps or 3D tracking videos, and the number of 3D tracking points.
To achieve this, we randomly selected 50 videos from the validation split of the DAVIS~\cite{Pont-Tuset_arXiv_2017} and MiraData~\cite{ju2024miradatalargescalevideodataset} video dataset. We extract the first-frame images as the input image and apply different models to re-generate these videos.
To evaluate the quality of the generated videos, we compute PSNR, SSIM~\cite{1284395}, LPIPS~\cite{zhang2018unreasonableeffectivenessdeepfeatures}, and FVD~\cite{unterthiner2019accurategenerativemodelsvideo} between the generated videos and the ground-truth videos.

\subsubsection{Depth maps vs. 3D tracking videos}

\begin{figure}
    \centering
    \includegraphics[width=\linewidth]{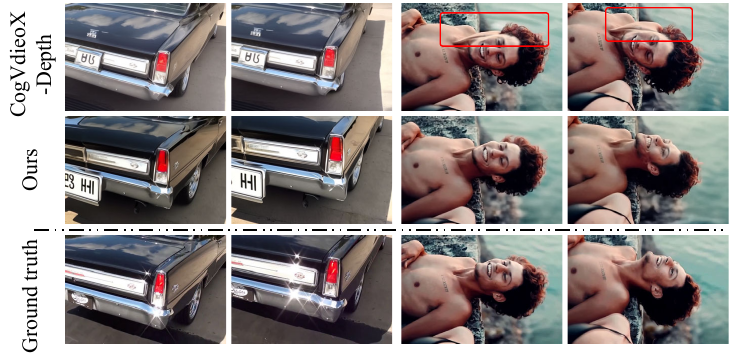}
    \caption{\textbf{Generated videos using depth maps or 3D tracking videos as control signals.} Our 3D tracking videos provide better quality on the cross-frame consistency for video generation than depth maps.}
    \vspace{-15pt}
    \label{fig:depth-tracking}
\end{figure}

To illustrate the effectiveness of our 3D tracking videos, we compare \methodname with a baseline using depth maps as conditions instead of 3D tracking videos. Specifically, the baseline adopts the same architecture as \methodname but replaces the 3D tracking video with a depth map video. We adopt the Depth Pro~\cite{bochkovskii2024depth} to generate the video depth video for this baseline method. As shown in \autoref{tab:ablation}, our model outperforms this baseline in all metrics, demonstrating that the 3D tracking videos provide a better signal for the diffusion model to recover groud-truth videos than the depth map conditions. \autoref{fig:depth-tracking} shows the generated videos, which demonstrate that our method produces more consistent videos with the ground truth. The main reason is that the 3D tracking videos effectively associate different frames of a video while the depth maps only provide some cues of the scene structures without constraining the motion of the video.

\subsubsection{Point density}

In \autoref{tab:ablation}, we further present an ablation study with varying numbers of 3D tracking points as control signals. The number of 3D tracking points ranges from 900 (30$\times$30) to 8100 (90$\times$90). Though the generated videos with 4900 tracking points perform slightly better than the other ones, the visual qualities of 2500, 4900, and 8100 tracking points are very similar to each other. Since tracking too many points with SpatialTracker~\cite{xiao2024spatialtracker} would be slow, we choose 4900 as our default setting in all our other experiments using 3D point tracking. 


\subsubsection{Runtime}
In the inference stage, we employ the DDIM~\cite{song2020denoising} sampler with 50 steps, classifier-free guidance of magnitude 7.0, which costs about 2.5 minutes to generate 49 frames on a H800 GPU at a resolution of 480$\times$720.

\begin{figure}
    \centering
    \includegraphics[width=\linewidth]{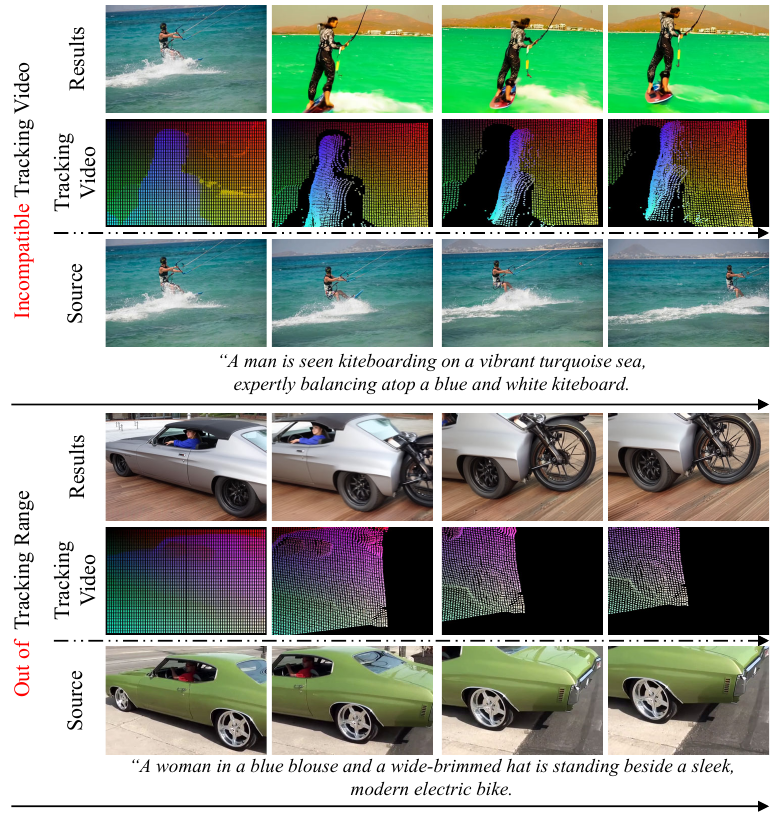}
    \caption{\textbf{Failure cases}. (Top) Incompatible tracking video. When a tracking video that does not correspond to the structures of the input image is provided, \methodname will generate a video with a scene transition to a compatible new scene.
    (Bottom) Out of tracking range. For regions without 3D tracking points, the tracking video fails to constrain these regions and \methodname may generate some uncontrolled content.} 
    \label{fig:tracking1}
\end{figure}

\section{Limitations and Conclusions}

\textbf{Limitations and future works}. 
Though \methodname achieves control over the video generation process in most cases, it still suffers from multiple failure cases mainly caused by incorrect 3Dtracking videos.
The first failure case is that the input image should be compatible with the 3D tracking videos. Otherwise, the generated videos would be implausible as shown in Figure~\ref{fig:tracking1} (top). Another failure case is that for regions without 3D tracking points, the generated contents may be out-of-control and produce some unnatural results (Figure~\ref{fig:tracking1} (bottom)).
For future works, we currently rely on provided animated meshes or existing videos to get high-quality 3D tracking videos and a promising direction is to learn to generate these 3D tracking videos with a new diffusion model.

\noindent\textbf{Conclusions}. In this paper, we introduce Diffusion as Shader (\methodname) for controllable video generation. 
The key idea of \methodname is to adopt the 3D tracking videos as 3D control signals for video generation. The 3D tracking videos are constructed from colored dynamic 3D points which represent the underlying 3D motion of the video. Then, diffusion models are applied to generate a video following the motion of the 3D tracking video. We demonstrate that the 3D tracking videos not only improve the temporal consistency of the generated videos but also enable versatile control of the video content, including mesh-to-video generation, camera control, motion transfer, and object manipulation.

\bibliographystyle{ACM-Reference-Format}
\bibliography{ref}

\end{document}